\documentclass[runningheads]{llncs}
\usepackage[T1]{fontenc}
\usepackage{float}
\usepackage{amssymb}
\usepackage{booktabs}
\usepackage{array}
\usepackage{multirow}
\usepackage{amsmath}
\usepackage{graphicx}
\usepackage{url}
\usepackage[misc,geometry]{ifsym}
\usepackage{color}

\begin{document}
\sloppy 
\title{DSAM-GN:Graph Network based on Dynamic Similarity Adjacency Matrices for Vehicle Re-identification
\thanks{This work was supported by the Open Research Fund of MOE Eng. Research Center of HW/SW Co-Design Tech. and App. , and the Science and Technology Commission of Shanghai Municipality (22DZ2229004).}}
\titlerunning{DSAM-GN}
%
\author{Yuejun Jiao\inst{1,2}\orcidID{0009-0007-7224-0192}
\and Song Qiu\inst{1,2}\textsuperscript{(\Letter)}
\and Mingsong Chen\inst{1}
\and Dingding Han\inst{3,4}
\and Qingli Li\inst{2}
\and Yue Lu\inst{2}}
\authorrunning{YJ Jiao et al.}
%
\institute{MOE Engineering Research Center of Software/Hardware Co-design Technology and Application, East China Normal University, Shanghai 200062, China\\
\inst{\textsuperscript{\Letter}} Corresponding author,
Email:\email{sqiu@ee.ecnu.edu.cn}\\
\and Shanghai Key Laboratory of Multidimensional Information Processing, East China Normal University, Shanghai 200241, China
\and School of Information Science and Technology, Fudan University, \\
Shanghai 200433, China
\and  Shanghai Artificial lntelligence Laboratory, Shanghai 200232, China}

%
\maketitle              
\begin{abstract}
In recent years, vehicle re-identification (Re-ID) has gained increasing importance in various applications such as assisted driving systems, traffic flow management, and vehicle tracking, due to the growth of intelligent transportation systems. However, the presence of extraneous background information and occlusions can interfere with the learning of discriminative features, leading to significant variations in the same vehicle image across different scenarios. This paper proposes a method, named graph network based on dynamic similarity adjacency matrices (DSAM-GN), which incorporates a novel approach for constructing adjacency matrices to capture spatial relationships of local features and reduce background noise. Specifically, the proposed method divides the extracted vehicle features into different patches as nodes within the graph network. A spatial attention-based similarity adjacency matrix generation (SASAMG) module is employed to compute similarity matrices of nodes, and a dynamic erasure operation is applied to disconnect nodes with low similarity, resulting in similarity adjacency matrices. Finally, the nodes and similarity adjacency matrices are fed into graph networks to extract more discriminative features for vehicle Re-ID. Experimental results on public datasets VeRi-776 and VehicleID demonstrate the effectiveness of the proposed method compared with recent works.
\keywords{Vehicle re-identification  \and Graph network \and Spatial attention.}
\end{abstract}

\section{Introduction}
Vehicle re-identification (Re-ID) is a task that aims to identify a target vehicle across video streams captured by different cameras. It has gained increasing importance in applications such as assisted driving systems, traffic flow management, and vehicle tracking within intelligent transportation systems. However, the presence of extraneous background information and occlusions can introduce interference and hinder the learning of discriminative features, resulting in significant feature variations of the same vehicle image in different scenarios. Therefore, it is crucial to remove extraneous information and minimize the interference of background noise in vehicle Re-ID tasks.

Various methods are proposed for fine-grained feature extraction to eliminate the interference of redundant information. These methods can be categorized into three aspects: knowledge-based methods\cite{huang2022KPGST,li2022DF-CVTC,zhang2022VGM}, uniform spatial division methods\cite{qian2020SAN,shen2022PFMN}, and part-level detection methods\cite{meng2020PVEN,Yu2023SOFCT}. Knowledge-based methods utilize metadata such as orientation, color, car type, key points, viewpoint, and spatiotemporal information to enhance the identification of vehicle details. Uniform spatial division methods divide the feature map horizontally or vertically into multiple parts and extract features separately from each part. Part-level detection methods employ image segmentation to semantically divide vehicles into multiple regions (e.g., roof, wheels, and windows) and extract features from these segmented regions. All the aforementioned methods  facilitate a comprehensive analysis of both the overall appearance and specific components of a vehicle, thereby enabling the extraction of intricate details. However, knowledge-based and part-level detection methods require additional annotations, the uniform spatial division method does not necessitate annotations but it is susceptible to partition misalignment. Additionally, the feature extraction methods employed in these approaches ignore the relationships among part regions.

In this paper, we propose a novel graph network based on dynamic similarity adjacency matrices (DSAM-GN) method for vehicle Re-ID. Our method aims to capture spatial relationships among local features and reduce background noise from the vicinity of vehicles. To achieve fine-grained feature extraction, the extracted vehicle features are divided into different patches. Unlike traditional CNN networks that overlook the correlation among local patches, we introduce a graph network to capture the spatial relationships among these patches. One challenge when applying graph networks to image representation is determining how to establish edges between nodes and which nodes to connect. In response to this challenge, we introduced a novel approach of utilizing the spatial attention mechanism to generate adjacency matrices. To overcome the issue of redundant background information, we employ a spatial attention-based similarity adjacency matrix generation (SASAMG) module to compute similarity matrices of patches. Furthermore, the SASAMG module employs dynamic erasure operation to disconnect nodes with low similarity, resulting in similarity adjacency matrices. Finally, the patches and similarity adjacency matrices are fed into graph networks to extract discriminative features for vehicle Re-ID.

The main contributions of this paper are as follows:
\begin{itemize}
\item We propose a novel graph network based on dynamic similarity adjacency matrices (DSAM-GN) method that combines a spatial attention mechanism to propose a new approach for constructing adjacency matrices required for the graph network. This method effectively captures spatial relationships among local features and reduces background noise from the vicinity of vehicles without any additional annotations.
\item We design a spatial attention-based similarity adjacency matrix generation (SASAMG) module, which employs a spatial attention mechanism and dynamic erasure operation to optimize connections between nodes and generate a similarity adjacency matrix. By erasing attention on nodes with background noise, this module establishes a fundamental basis for the learning of discriminative features.
\item Extensive experiments on public datasets VeRi-776\cite{liu2016veri776_1,liu2016veri776_2,liu2017veri776_3} and VehicleID\cite{liu2016vehicleID} demonstrate the effectiveness of our proposed method compared with recent works.
\end{itemize}

\section{Related Work}
\subsection{CNNs and Graph Networks}
Deep learning techniques are widely adopted in vehicle Re-ID methods. Convolutional neural networks (CNNs) emerge as the dominant approach for deep feature extraction due to their exceptional capability to capture discriminative features. Several works\cite{huang2022KPGST,shen2022PFMN} employ CNN architectures as feature extractors, enabling the learning of both global and local features. However, CNNs often focus only on local information and fail to capture the relationships among different regions with intricate local details. To address this limitation, graph networks (GNs) are introduced as a viable solution, allowing the exploration of interconnections among local features derived from different regions. The graph convolutional network (GCN) \cite{kipf2016GCN} updates node representations by aggregating information from neighboring nodes, enabling the node representation to inherit information from nearby regions. The graph attention network (GAT) \cite{velickovic2018gat} utilizes attention mechanism to control the influence of different neighboring nodes on the target node representation, thus reducing the impact of irrelevant nodes. GNs have been successfully applied in various domains, including computer vision\cite{guo2018neural,taufique2021labnet,xu2021HSS-GCN,zhu2020SGAT,liu2020PCRNet}, social networks\cite{bian2020rumor}, and recommendation systems\cite{chen2020revisiting}.

\subsection{Node and Edge Construction in GNs for Vehicle Re-ID}
When applying graph networks to image representation, careful consideration must be given to node definition and edge construction. The local graph aggregation network with class balanced loss (LABNet)\cite{taufique2021labnet} defines spatial regions of the feature map as nodes and establishes edges among nodes using a simple 8-neighborhood connectivity approach. This straightforward method can introduce redundant background information, which negatively impacts model performance. The hierarchical spatial structural graph convolutional network (HSS-GCN)\cite{xu2021HSS-GCN} uniformly divides the global feature map into five regions: upper-left, upper-right, middle, down-left, and down-right, and treats each of these regions as a node in a graph. Edges are formed among these regions and a global node. These regions still contain background noise. The structured graph attention network (SGAT)\cite{zhu2020SGAT} creates nodes based on 20 selected landmarks detected by a landmark detection module, and edges among the landmarks are determined by their Euclidean distances being smaller than a predefined threshold. This method relies on expensive additional annotations for landmark detection. The parsing-guided cross-part reasoning network (PCRNet)\cite{liu2020PCRNet} employs part-level segmentation to divide vehicles into regions, constructing a part-neighboring graph using regional features. The part-level segmentation approach also requires costly additional annotations. In this paper, the graph network based on dynamic similarity adjacency matrices (DSAM-GN) divides the extracted vehicle features into different patches as nodes. A spatial attention mechanism and dynamic erasure operation are applied to optimize connections between nodes.
\section{Proposed Method}
\begin{figure}
\includegraphics[width=\textwidth]{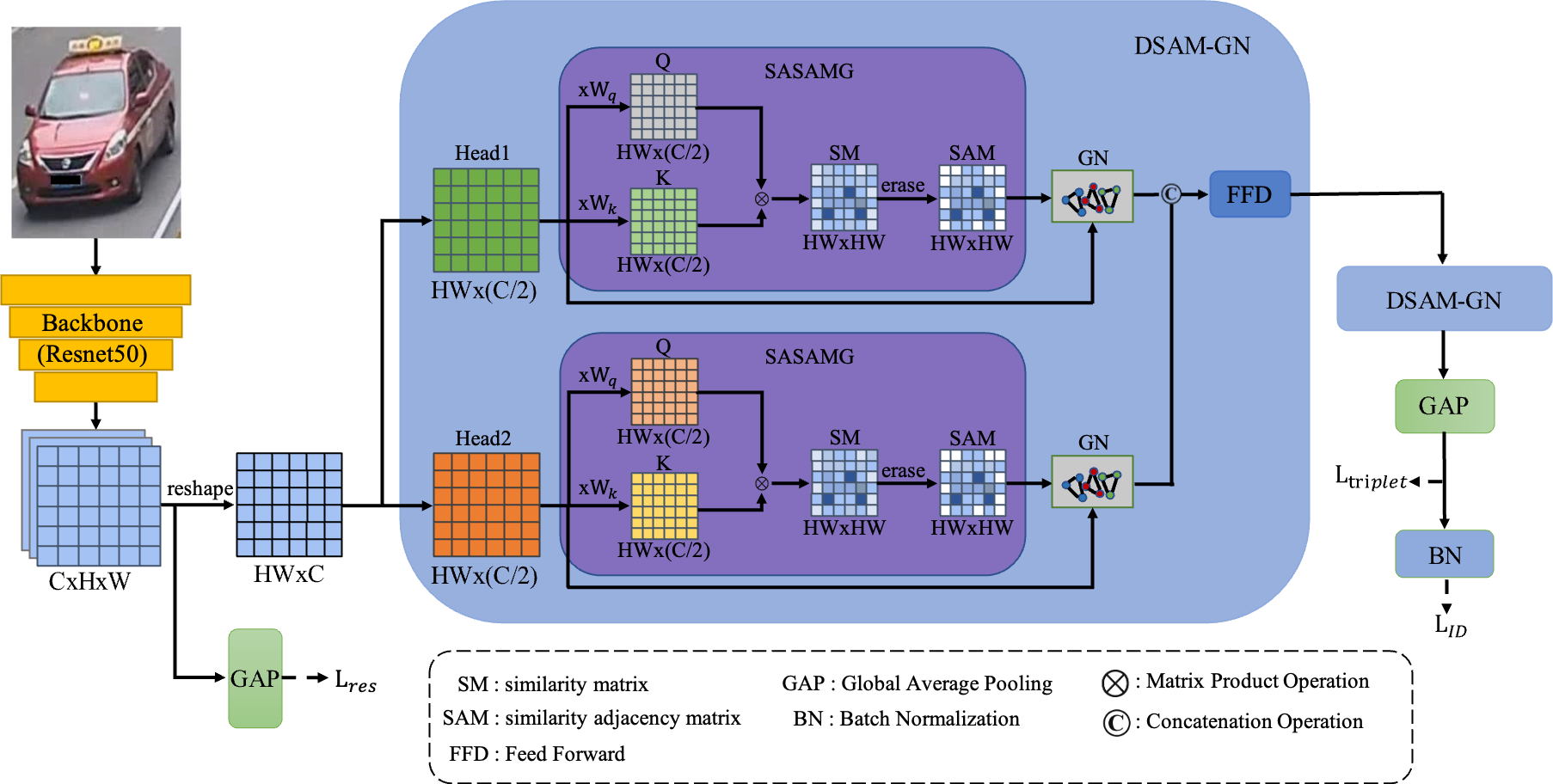}
\caption{The overall architecture of the proposed DSAM-GN model.} \label{model}
\end{figure}
\subsection{Overview}
Fig.\ref{model} illustrates the proposed model's architecture. The backbone network initially processes the input image to extract fundamental vehicle features, which are divided into multiple patches before inputting into the DSAM-GN module for feature extraction. Within SASAMG, the input features are multiplied by two trainable parameter matrices, $W_q$ and $W_k$, to produce the query and key matrices, respectively. The matrix product of the query and key matrices is then computed, and the resulting values are softmax-normalized to obtain the similarity matrix(SM). Then, connections between patches with low similarity are erased to obtain a similarity adjacency matrix(SAM). Each patch is treated as a node, and both the nodes and the SAM are fed into the graph network(GN), which captures feature relationships among the nodes. The Feed Forward(FFD) module consists of a multi-layer perceptron and ReLU activation function, which are used to aggregate the features extracted from the preceding two branches. The motivation for adopting two branches arises from the outstanding performance of the multi-head attention mechanism in Transformer\cite{vaswani2017Transformer}. The features are then processed through the second DSAM-GN module and undergo global average pooling (GAP) and batch normalization (BN) to produce the final output. Importantly, GAP replaces the fully connected layer, significantly reducing the number of network parameters and preventing model overfitting.

\subsection{DSAM-GN}
The DSAM-GN module explores the relationships among different patches while discarding redundant patches. The backbone network extracts features from the image, serving as the original appearance representation. The resulting feature is represented by a $C \times H \times W$ tensor, where $C$, $H$, and $W$ indicate the number of feature channels, height, and width, respectively. The feature is reshaped into $N \times C$ for the subsequent similarity evaluation, with $N$ representing the number of patches, and $N = H \times W$. The embedding features of the patches are formed as shown in Eq.(\ref{eq:1}):

\begin{equation}
X^{input}=\left[X_{1},X_{2},\dots,X_{i},\dots,X_{N}\right]+P_{pos}
\label{eq:1}
\end{equation}

Here, $X^{input}\in\mathbb{R}^{N\times C}$ represents the input to the DSAM-GN module, and $P_{pos}\in\mathbb{R}^{N \times C}$ denotes the learnable positional encoding. The input features $X^{input}$ are split along the channel dimension into two branches and fed into the SASAMG modules, respectively. Inside the SASAMG module, the input features are linearly transformed into queries $Q\in\mathbb{R}^{N\times (C/2)}$ and keys $K\in\mathbb{R}^{N \times (C/2)}$. The matrix product of them is applied to calculate the similarity matrix $S$, which represents the similarity among the patches according to Eq.(\ref{eq:2}):

\begin{equation}
S(Q,K)=\text{softmax}\left(\frac{Q K^{T}}{\sqrt{d_{k}}}\right)
\label{eq:2}
\end{equation}

Furthermore, to obtain a similarity adjacency matrix, a dynamic erasure operation is applied to disconnect patches with low similarity. The similarity adjacency matrix $A$ is computed as shown in Eq.(\ref{eq:3}):

\begin{equation}
A_{i,j}=
\begin{cases}  
S_{i,j}& S_{i,j}>p(S, \beta)\\
0& S_{i,j} \leq p(S, \beta)
\end{cases}
\label{eq:3}
\end{equation}

Here, the function $p$ calculates the percentile value of the similarity matrix $S$, with the hyperparameter $\beta \in [0,100]$ representing the percentile index for patches with low similarity. Specifically, we first flatten $S$ into a one-dimensional array $B$, and sort its elements in ascending order. Let $n=N \times N$ be the length of $B$. For instance, if we want to calculate the 85 percentile, $p(S, 85) = B_k$, where $k = \lceil 85\% n \rceil$.

Each patch is treated as a node, and both the nodes and the similarity adjacency matrix are fed into the graph network, which captures feature relationships among the nodes. The computation of the graph network is expressed by Eq.(\ref{eq:4}):

\begin{equation}
\mathbf{h}'_i = \sigma\left(\sum_{j \in \mathcal{N}_i} A_{i,j} \mathbf{W}\mathbf{h}_j\right)
\label{eq:4}
\end{equation}

Here, $\mathbf{h}_j$ represents the feature vector of the $i$-th node's neighbor. The weight matrix $\mathbf{W}$ is a trainable parameter, and $\mathcal{N}_i$ denotes the set of neighboring nodes for the $i$-th node, which represents the set of nodes that remain connected to the $i$-th node after dynamic erasure operation. $\sigma$ represents the activation function, and in this paper, the ReLU activation function is used. Finally, the output is the representation vectors of all nodes: $\mathbf{h}'_1, \mathbf{h}'_2, \dots, \mathbf{h}'_i, \dots, \mathbf{h}'_N$.

\subsection{Loss Function}
As depicted in Fig.\ref{model}, we adopt a multi-task learning approach for joint training. The output of the GAP layer (after the DSAM-GN module) is used to compute the triplet loss ($L_{\text{triplet}}$), while the output of the BN layer is employed to calculate the ID loss ($L_{\text{ID}}$), which is a cross-entropy loss. Moreover, to address the issue of large intra-class distance and small inter-class distance, we incorporate the triplet loss ($L_{\text{res}}$) as an auxiliary supervision for the backbone network's output. Three hyperparameters ($\alpha, \beta, \gamma$) correspond to the coefficients of the aforementioned three loss functions. In order to avoid excessive fine-tuning of hyperparameters, we set $\alpha = \beta = \gamma = 1$ in the following experiments. Thus, the total loss function of the proposed method can be formulated as shown in Eq.(\ref{eq:5}):

\begin{equation}
L_{total}=\alpha L_{res}+\beta L_{triplet}+\gamma L_{ID}
\label{eq:5}
\end{equation}

\section{Experiments}
In this section, we present the experimental results and analysis of our proposed model for vehicle Re-ID. We evaluate the model on the VeRi-776\cite{liu2016veri776_1,liu2016veri776_2,liu2017veri776_3} and VehicleID\cite{liu2016vehicleID} datasets, compare its performance with state-of-the-art methods, and conduct an ablation study to assess the effectiveness of our proposed method.

\subsection{Implementation Details}
Before training, we randomly applied crop, flip, and pad operations on the images with a certain probability. The images were then uniformly resized to 256 x 256 pixels. To construct our network, we employed a ResNet50 as the backbone architecture. The ResNet50 was initialized with pre-trained weights from the ImageNet dataset. The training parameters varied between the VeRi-776 and VehicleID datasets due to differences in image quality, quantity, and perspective. The VeRi-776 dataset was trained using one GPU with a batch size of 128 and SGD optimization with warm-up strategy. The learning rate increased to 0.01 after 3000 iterations and gradually decreased using cosine annealing until the 60th epoch. On the other hand, the VehicleID dataset was trained using two GPUs with a batch size of 256 and Adam optimization with warm-up strategy. The learning rate was initially set to 0.000035 and increased to 0.0002 after 2000 iterations. Then, it was reduced by a factor of 0.1 at the 30th, 70th, and 90th epochs, culminating in a total of 100 epochs. The model was implemented using the PyTorch framework and trained and tested on an NVIDIA RTX 3090.

\subsection{Experimental Results and Analysis}

\subsubsection{Results on VeRi-776 Dataset}
We first evaluated our model on the VeRi-776 dataset and compared its performance against various state-of-the-art methods. Table \ref{table1} presents the comparison results. Our model achieved state-of-the-art performance compared with other methods. Specifically, our method achieved a higher mAP score, surpassing the baseline by 1.12\%. Moreover, the Rank-1 and Rank-5 scores showed improvements of 0.42\% and 0.40\%, respectively, over the baseline. These results provide strong evidence for the effectiveness of our model on the VeRi-776 dataset.

\begin{table}[htbp]
\centering
\setlength{\extrarowheight}{2pt}
\caption{Comparison with state-of-the-art results(\%) on VeRi-776.The best result is bolded.}\label{table1}
\begin{tabular}{c|c|c|c|c}
\hline
\multicolumn{1}{c|}{\multirow{2}{*}{Method}} & \multicolumn{1}{c|}{\multirow{2}{*}{Publicaiton}} & \multicolumn{3}{c}{VeRi-776}       \\ \cline{3-5} 
 &   & mAP & Rank-1 & Rank-5 \\ \hline
HSS-GCN\cite{xu2021HSS-GCN}       & ICPR'21     & 44.80  & 64.40  & 86.10             \\
DF-CVTC\cite{li2022DF-CVTC}       & TETCI'22       & 61.06    &  91.36     &  95.77            \\
SGAT\cite{zhu2020SGAT}            & ACMMM'20  & 65.66  & 89.69  & -          \\
KPGST\cite{huang2022KPGST}        & Electronics'22    &  68.73   & 92.35  &  93.92  \\
SAN\cite{qian2020SAN}            & MST'20     & 72.50    &  93.30  & 97.10  \\
VGM\cite{zhang2022VGM}            & APIN’22    & 73.32    & 92.82      & 95.21  \\
PCRNet\cite{liu2020PCRNet}       &  ACMMM'20  & 78.60  & 95.40   & 98.40             \\
PVEN\cite{meng2020PVEN}          & CVPR'20     & 79.50    &  95.60     &  98.40            \\
LABNet\cite{taufique2021labnet}   &  Neurocput'21    & 79.50  & 95.70  &  -            \\
SOFCT\cite{Yu2023SOFCT}          &  TITS'23    & 80.70    &  96.60     &  98.80            \\
PFMN\cite{shen2022PFMN}           & CIS'22     & 81.20    &  96.80     & 97.60  \\
MRF-SAPL\cite{pang2023MRF-SAPL}   & Entropy'23  & 81.50  & 94.70  &  98.70             \\
\hline
baseline            &  -    & 81.09    &  96.72     &    98.33        \\
DSAM-GN(ours)       &  -   & {\bfseries 82.22} & {\bfseries 97.38} & {\bfseries 98.75}  \\ \hline
\end{tabular}
\end{table}

\subsubsection{Results on VehicleID Dataset}
Next, we conducted experiments on the VehicleID dataset and compared our results against state-of-the-art methods, which are presented in Table \ref{table2}. Our model outperformed the baseline in terms of the mAP, Rank-1, and Rank-5 metrics. Specifically, our approach  achieved the best mAP scores across all three VehicleID subsets (800, 1600, and 2400), outperforming other methods. This indicated that DSAM-GN was effective at identifying vehicles in terms of average precision. DSAM-GN has demonstrated strong performance in both Rank-1 and Rank-5 scores, with its performance being surpassed only by PCRNet. Notably, DSAM-GN achieves the best Rank-1 score on the VehicleID-2400 subset, outperforming PCRNet. Although PCRNet achieved excellent scores, its use of segmentation techniques required expensive annotation. However, DSAM-GN's results were still impressive, considering that it did not rely on costly annotations.

\begin{table}[htbp]
\centering
\setlength{\extrarowheight}{5pt}
\caption{Comparison with state-of-the-art results(\%) on VehicleID.The best result is bolded.}\label{table2}
\resizebox{\columnwidth}{!}{\begin{tabular}{c|c|ccc|ccc|ccc}
\hline
\multicolumn{1}{c|}{\multirow{2}{*}{Method}} & \multicolumn{1}{c|}{\multirow{2}{*}{Publicaiton}}  & \multicolumn{3}{c|}{VehicleID-800}    & \multicolumn{3}{c|}{VehicleID-1600}      & \multicolumn{3}{c}{VehicleID-2400}            \\ \cline{3-11} 
  &      & mAP & Rank-1 & Rank-5 & mAP & Rank-1 & Rank-5 & mAP & Rank-1 & Rank-5 \\ \hline
HSS-GCN\cite{xu2021HSS-GCN}      & ICPR'21  & 77.30  & 72.70  & 91.80  & 72.40  & 67.90  & 87.80  & 66.10  & 62.40  & 84.30       \\
DF-CVTC\cite{li2022DF-CVTC}      & TETCI'22  & 78.03 &  75.23 &  88.11 &  74.87 & 72.15 &  84.37 &  73.15 &  70.46 &  82.13      \\
SGAT\cite{zhu2020SGAT}           & ACMMM'20  & 81.49  & 78.12  & -  & 77.46    & 73.98  & -  & 75.35  & 71.87  & -       \\
SAN\cite{qian2020SAN}            & MST'20    &  -     &  79.70  &  94.30 &  -     &  78.40  &  91.30  &  -     &  75.60  & 88.30   \\
LABNet\cite{taufique2021labnet}  & Neurocput'21  & 87.50  &  81.20  &  -     &  84.20 & 78.00   &  -     &  80.80  &  73.50  &  -       \\
MRF-SAPL\cite{pang2023MRF-SAPL}  & Entropy'23  & -  &  84.30  &  97.70  &  -  &  79.60  &  94.10  &  -     &  76.30  &  91.60      \\
SOFCT\cite{Yu2023SOFCT}          &  TITS'23    & 89.80  &  84.50  &  96.80  &  86.40  &  80.90  &  95.20  &  84.3     &  78.70  &  93.70      \\
PVEN\cite{meng2020PVEN}          & CVPR'20        &  -    &  84.70  &  97.00  &  -    & 80.60   &  94.50  &  -     &  77.80  &  92.00       \\
PFMN\cite{shen2022PFMN}          & CIS'22     & -   &  85.60 &  96.80 &  -     &  81.40  &  94.10  &   -   &  80.00    &  92.00    \\
PCRNet\cite{liu2020PCRNet}       &  ACMMM'20      &  -    &  {\bfseries86.60}  &  {\bfseries98.10}  &  -    & {\bfseries82.20}   &  {\bfseries96.30}  &  -     &  80.40  &  {\bfseries94.20}  \\ \hline
baseline   &   -         &  75.89   &  66.33  &  89.38   &   69.20    &   58.72    &    82.55    &        64.41   &    53.77   &    76.83     \\
DSAM-GN(ours)       &   -        &  {\bfseries90.42}   &  85.63  &  96.96   &   {\bfseries86.60}    &   81.62   & 95.22   &    {\bfseries84.66}   &    {\bfseries81.26}   &   93.89  \\  \hline
\end{tabular}}
\end{table}

\begin{figure}
\includegraphics[width=\textwidth]{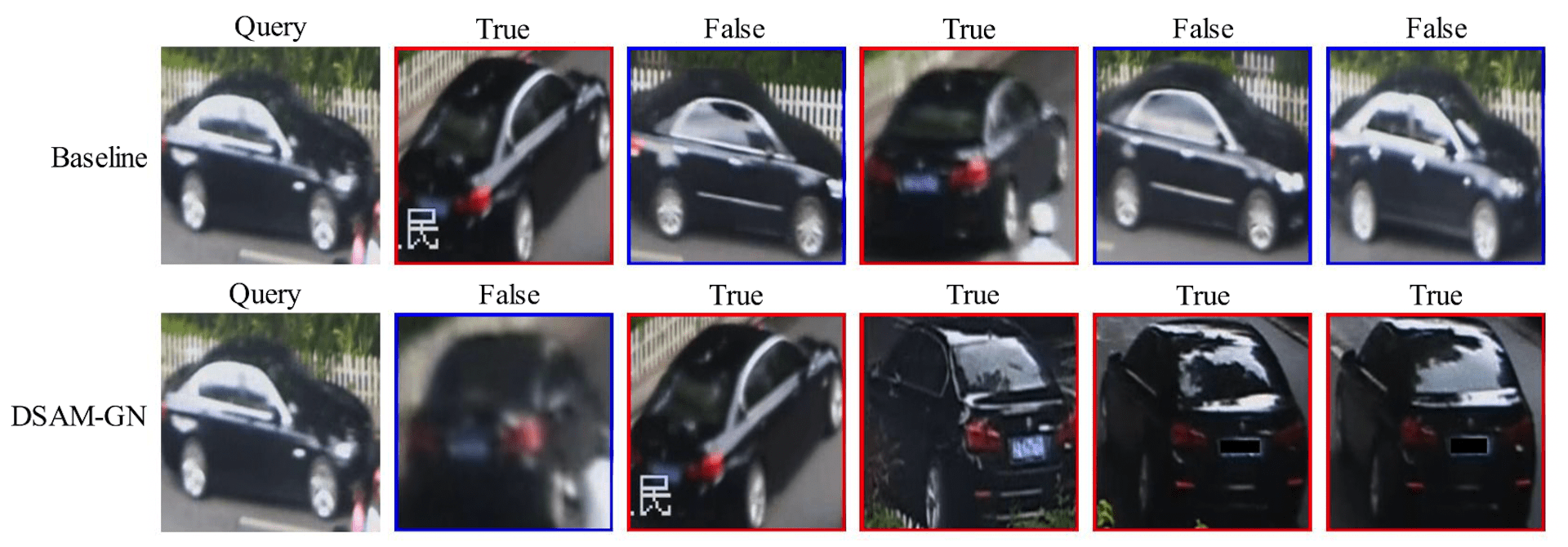}
\caption{Rank-5 visualization examples on VeRi-776.} \label{result}
\end{figure}

\subsubsection{Visualization}
To visually assess the performance of our proposed model, we present the rank-5 retrieval results of an example query image from the VeRi-776 dataset, as illustrated in Figure \ref{result}. The top five images retrieved by the Baseline and DSAM-GN approaches are displayed in the first and second rows, respectively. Correct retrieval results are indicated by red boxes, while incorrect results are highlighted with blue boxes. It is evident that our proposed model outperforms the baseline and exhibits superior ability in distinguishing similar vehicles.

\begin{figure}
\includegraphics[width=\textwidth]{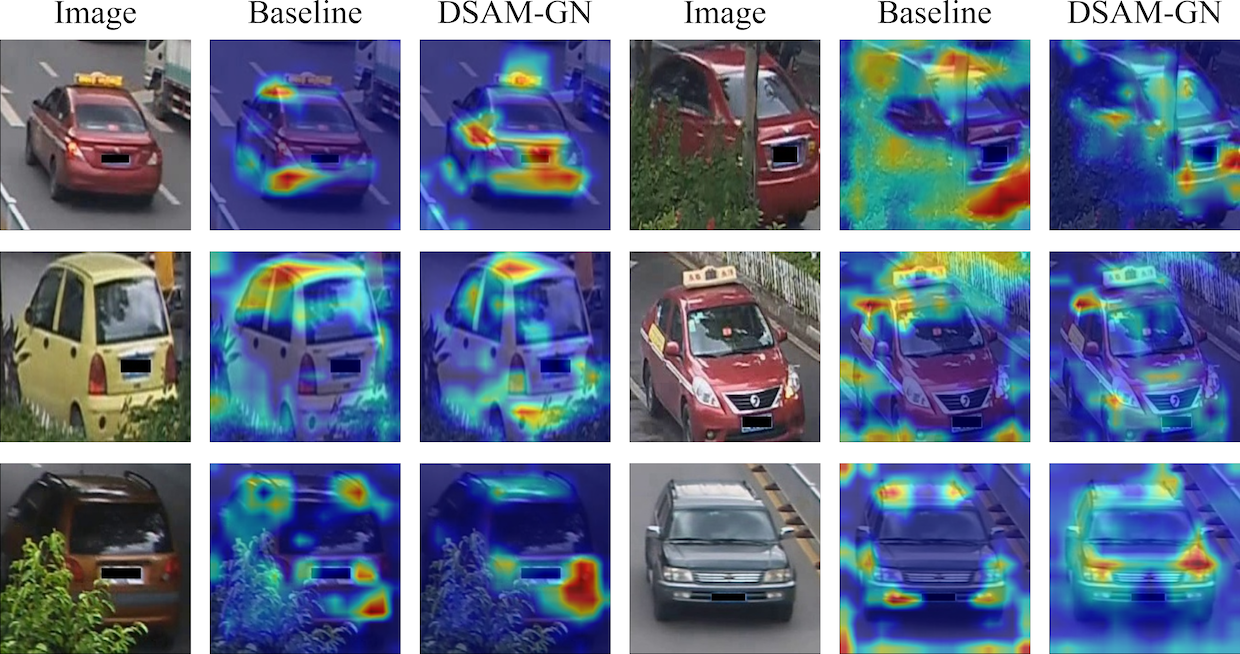}
\caption{Gradient-weighted Class Activation Mapping(Grad-CAM) visualization of attention maps.} \label{heatmap}
\end{figure}

Additionally, we utilized Grad-CAM \cite{selvaraju2017grad-cam} to generate attention maps for challenging samples with background occlusions. Fig.\ref{heatmap} shows the attention maps, illustrating the model's focus. The attention maps of the baseline model revealed a strong emphasis on the background information, negatively impacting its performance. In contrast, our proposed model exhibited a stronger focus on the vehicles themselves, effectively recalibrating the model's attention and reducing extraneous focus on the background. This visualization provides qualitative evidence of the superiority of our approach in handling background occlusions and improving the model's discriminative ability.

\begin{figure}
\includegraphics[width=\textwidth]{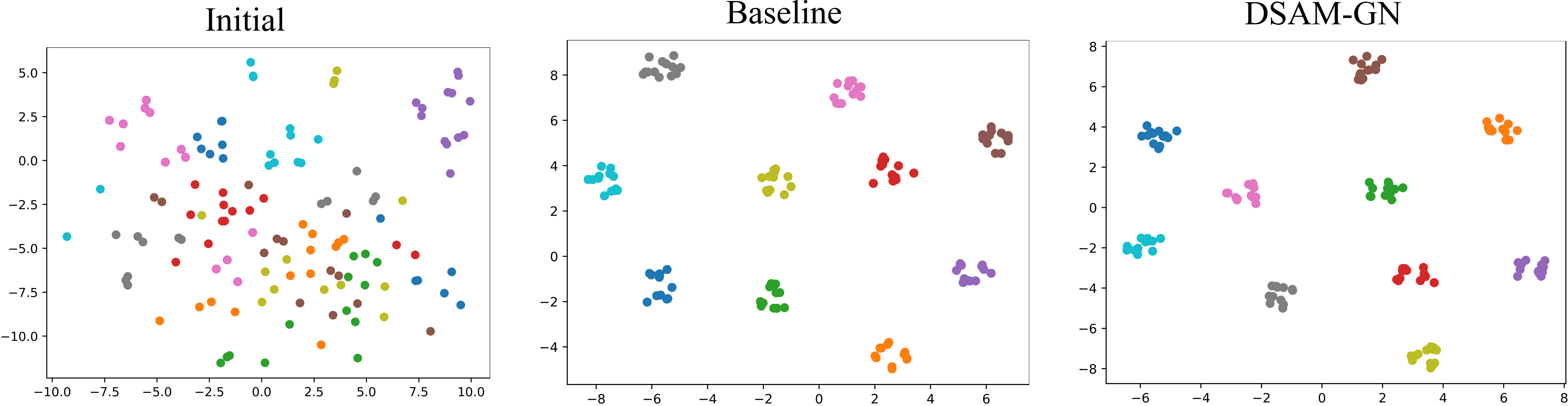}
\caption{t-SNE visualization of the learned feature space.} \label{scatter}
\end{figure}

Furthermore, we employed t-SNE\cite{van2008t-sne} to visualize scatter plots that visualize the distribution of data points in the feature space before and after model training. Fig.\ref{scatter} shows these scatter plots. The scatter plot before training exhibited a disordered distribution of distinct categories. Although the baseline model facilitated clustering of different categories, it is evident from the vertical axis ($-4, 8$) that it still had shortcomings, such as a small inter-class distance. In contrast, our proposed model's performance on the vertical axis ($-8, 8$) demonstrated a significant improvement in inter-class distance. This observation highlights that our model can effectively learn discriminative features and improve the separation of different vehicle categories.

\subsection{Ablation Study}
To further evaluate the effectiveness and robustness of our proposed method, we conducted an ablation study on the VeRi-776 and VehicleID datasets.

\subsubsection{Effectiveness of DSAM-GN}
In this ablation study, we evaluated the impact of different percentile  ($\beta$) for the DSAM-GN module on the VehicleID and VeRi-776 datasets. Table \ref{table3} presents the performance results. We can observed that the performance consistently outperformed the baseline for different percentile values. Notably, the best performance was achieved when $\beta$ was set to 95, resulting in an improvement of mAP, Rank-1, and Rank-5 scores over the baseline. These results demonstrate the effectiveness of the DSAM-GN module in reducing background noise.

\begin{table}[htbp]
\centering
\setlength{\extrarowheight}{5pt}
\caption{Evaluation of the impact(\%) of percentile $\beta$ for DSAM-GN on VehicleID and VeRi-776. the best result is bolded.}
\label{table3}
\resizebox{\columnwidth}{!}{\begin{tabular}{c|ccc|ccc|ccc|ccc}
\hline 
\multicolumn{1}{c|}{\multirow{2}{*}{Method}} & \multicolumn{3}{c|}{VehicleID-800}    & \multicolumn{3}{c|}{VehicleID-1600}      & \multicolumn{3}{c|}{VehicleID-2400}       & \multicolumn{3}{c}{VeRi-776}        \\ \cline{2-13} 
    & mAP & Rank-1 & Rank-5 & mAP & Rank-1 & Rank-5 & mAP & Rank-1 & Rank-5 & mAP & Rank-1 & Rank-5 \\ \hline
Baseline                &  75.89   &  66.33  &  89.38   &   69.20    &   58.72    &    82.55    &        64.41   &    53.77   &    76.83      & 81.09    &  96.72     &    98.33       \\ \hline 
DSAM-GN(0 percentile)     &  88.64   &  83.22  &  96.13   &   84.19    &   78.08    &    92.54    &        83.06   &    77.36   &    90.39      &    81.15      &  97.08   &  98.63     \\
DSAM-GN(75 percentile)     &  88.64   &  83.22  &  96.13   &   84.19    &   78.08    &    92.54    &        83.06   &    77.36   &    90.39     &    81.50      &  97.02   &  98.45     \\
DSAM-GN(85 percentile)     &  90.26   &  85.49  &  96.36   &   85.74    &   79.91    &    93.40    &        83.76   &    78.06   &    91.19     &    81.95      &  97.26   &  98.63     \\
DSAM-GN(95 percentile)      &  {\bfseries90.42}   &  {\bfseries85.63}  &  {\bfseries96.96}   &   {\bfseries86.60}    &   {\bfseries81.62}    &    {\bfseries95.22}    &    {\bfseries84.66}   &    {\bfseries81.26}   &    {\bfseries93.89}   & {\bfseries 82.22} & {\bfseries 97.38} & {\bfseries 98.75}   \\ 
DSAM-GN(98 percentile)     &  87.85   &  81.89  &  95.78   &   83.73    &   78.02    &    91.56    &        81.70   &    75.90   &    89.03    &    81.42      &  96.90   &  98.57   \\
\hline
\end{tabular}}
\end{table}

\section{Conclusion}
In this paper, we propose a novel graph network based on dynamic similarity adjacency matrices (DSAM-GN) method that combines a spatial attention mechanism to propose a new approach for constructing adjacency matrices required for the graph network. This method effectively captures spatial relationships among local features and reduces background noise without any additional annotations. We design a spatial attention-based similarity adjacency matrix generation (SASAMG) module, which employs a spatial attention mechanism and dynamic erasure operation to optimize connections between nodes and generate a similarity adjacency matrix. By erasing attention on nodes with background noise, this module establishes the foundation for learning discriminative local features. Extensive experiments on the VeRi-776 and VehicleID datasets demonstrated the effectiveness of our proposed method. Visual comparisons with the baseline model showcased that our method is more focused on the vehicles themselves and demonstrated a significant improvement in inter-class distance. These results highlight the potential of our approach for vehicle re-identification tasks.

%
%

\bibliography{paper} 
\bibliographystyle{splncs04}

\end{document}